\documentclass[conference]{IEEEtran}

\usepackage{cite}
\usepackage{amsmath,amssymb,amsfonts}
\usepackage{algorithmic}
\usepackage{algorithm}
\usepackage{graphicx}
\usepackage{textcomp}
\usepackage{xcolor}
\usepackage{url}
\usepackage[keeplastbox]{flushend}

\begin{document}
\title{AGBoost: Attention-based Modification of Gradient Boosting Machine}
\date{}

\author{
\IEEEauthorblockN{Andrei Konstantinov, Lev Utkin, Stanislav Kirpichenko}
\IEEEauthorblockA{Peter the Great St.Petersburg Polytechnic University \\ St.Petersburg, Russia 
\\ andrue.konst@gmail.com, lev.utkin@gmail.com, kirpichenko.sr@gmail.com}}

\maketitle

\begin{abstract}
A new attention-based model for the gradient boosting machine (GBM) called
AGBoost (the attention-based gradient boosting) is proposed for solving
regression problems. The main idea behind the proposed AGBoost model is to
assign attention weights with trainable parameters to iterations of GBM under
condition that decision trees are base learners in GBM. Attention weights are
determined by applying properties of decision trees and by using the Huber's
contamination model which provides an interesting linear dependence between
trainable parameters of the attention and the attention weights. This
peculiarity allows us to train the attention weights by solving the standard
quadratic optimization problem with linear constraints. The attention weights
also depend on the discount factor as a tuning parameter, which determines how
much the impact of the weight is decreased with the number of iterations.
Numerical experiments performed for two types of base learners, original
decision trees and extremely randomized trees with various regression datasets
illustrate the proposed model.
\end{abstract}

\section{Introduction}

One of the promising tools in deep learning is the attention mechanism which
assigns weight to instances or features in accordance with their importance
for enhancing the regression and classification performance. The attention
mechanism comes from the biological nature of the human perception to be
concentrated on some important parts of images, text, data, etc.
\cite{Niu-Zhong-Yu-21}. Following this property, various models of attention
have been developed in order to improve machine learning models. Many
interesting surveys devoted to different forms of the attention mechanism,
including transformers as the powerful neural network models, can be found in
\cite{Chaudhari-etal-2019,Correia-Colombini-21a,Correia-Colombini-21,Lin-Wang-etal-21,Niu-Zhong-Yu-21}%
.

An important peculiarity of the attention mechanism is that it is trainable,
i.e., it, as a model, contains trainable parameters. Due to this property most
attention models are components of neural networks \cite{Chaudhari-etal-2019},
and the attention trainable parameters are learned by using the gradient-based
algorithms which may lead to overfitting, expensive computations, i.e., the
attention models have the same problems as neural networks. In order to
overcome this difficulty and simultaneously to get attention-based models with
a simple training algorithm, Utkin and Konstantinov
\cite{Utkin-Konstantinov-22a} proposed a new model which is called the
attention-based random forest. According to this model, the attention weights
are assigned to decision trees in the random forest \cite{Breiman-2001} in a
specific way. Moreover, the attention weights have trainable parameters which
are learned on the corresponding dataset. One of the main ideas behind the
attention-based random forest is to apply the Huber's $\epsilon$-contamination
model \cite{Huber81} which establishes relationship between the attention
weights and trainable parameters. Various numerical examples with well-known
regression and classification datasets demonstrated outperforming results.

The random forest is a powerful ensemble-based model which is especially
efficient when we deal with tabular data. However, there is another
ensemble-based model, the well-known gradient boosting machine (GBM)
\cite{Friedman-2001,Friedman-2002}, which is reputed a more efficient model
for many datasets and more flexible one. GBMs have illustrated their
efficiency for solving regression problems \cite{Buhlmann-Hothorn-07}.

Following the attention-based random forest model, we aim to apply some ideas
behind this model to the GBM and to develop quite a new model called the
attention-based gradient boosting machine (AGBoost). In accordance with
AGBoost, we assign weights to each iteration of the GBM in a specific way
taking into account the tree predictions and the discount factor which
determines how much the impact of the attention weight is decreased with the
number of iterations. It is important to note that the attention mechanism
\cite{Zhang2021dive} was originally represented in the form of the
Nadaraya-Watson kernel regression model \cite{Nadaraya-1964,Watson-1964},
where attention weights conform with relevance of a training instance to a
target feature vector. The idea behind AGBoost is to incorporate the
Nadaraya-Watson kernel regression model into the GBM. We also apply the
Huber's $\epsilon$-contamination model where the contamination distribution
over all iterations is regarded as a trainable parameter vector. The training
process of attention weights is reduced to solving the standard quadratic
optimization problem with linear constraints. We consider AGBoost only for
solving regression problems. However, the results can be simply extended to
classification problems.

Numerical experiments with regression datasets are provided for studying the
proposed attention-based model. Two types of decision trees are used in
experiments: original decision trees and Extremely Randomized Trees (ERT)
proposed in \cite{Geurts-etal-06}. At each node, the ERT algorithm chooses a
split point randomly for each feature and then selects the best split among these.

The paper is organized as follows. Related work can be found in Section 2. A
brief introduction to the attention mechanism is given in Section 3. The
proposed AGBoost model for regression is presented in Section 4. Numerical
experiments illustrating regression problems are provided in Section 5.
Concluding remarks can be found in Section 6.

\section{Related work}

\textbf{Attention mechanism}. The attention mechanism is considered a powerful
and perspective tool for constructing machine learning models having accurate
performance in several applications. As a result, many classification and
regression algorithms have been added by attention-based models to improve
their performance. Attention models became the key modules of Transformers
\cite{Lin-Wang-etal-21}, which have achieved great success in many
applications and fields, including natural language processing and computer
vision. Surveys of various attention-based models can be found in
\cite{Chaudhari-etal-2019,Correia-Colombini-21a,Correia-Colombini-21,Liu-Huang-etal-21,Niu-Zhong-Yu-21}%
. In spite of efficiency of the attention models, they require to train the
softmax functions with trainable parameters that leads to computational
problems. Several methods of the softmax function linearization were developed
\cite{Choromanski-etal-21,Ma-Kong-etal-21,Peng-Pappas-etal-21,Schlag-etal-2021}%
.

Our aim is to propose the attention-based GBM modification which train the
attention parameters by means of the quadratic optimization that is simply solved.

\textbf{Gradient boosting machines}. The GBM is one of the most efficient tool
for solving regression and classification problems especially with tabular
data. Moreover, it can cope with non-linear dependencies
\cite{Natekin-Knoll-13}. Decision trees are often used in GBMs as basic
models. The original GBM \cite{Friedman-2001} is based on decision trees which
are sequentially trained to approximate negative gradients. Due to the success
of GBMs, various modifications have been developed, for example, the
well-known XGBoost \cite{Chen-Guestrin-2016}, pGBRT \cite{Tyree-etal-11}, SGB
\cite{Friedman-2002}. Advantages of decision trees in GBMs led to a
modification of the GBM on the basis of the deep forests
\cite{Zhou-Feng-2017a}, which is called the multi-layered gradient boosting
decision tree model \cite{Feng-Yu-Zhou-2018}. An interesting modification is
the soft GBM \cite{Feng-etal-20}. Another direction for modifying GBMs is to
use extremely randomized trees \cite{Geurts-etal-06} which illustrated the
substantial improvement of the GBM performance. Several GBM models have been
implemented by using modifications of extremely randomized trees and their
modifications \cite{Konstantinov-Utkin-20}.

We modify the GBM to incorporate the attention mechanism into the iteration
process and to weigh each iteration with respect to its importance in the
final prediction.

\section{Preliminary}

\subsection{The attention mechanism}

The attention mechanism can be viewed as a trainable mask which emphasizes
relevant information in a feature map. One of the clear explanations of the
attention mechanism is to consider it from the statistics point of view
\cite{Chaudhari-etal-2019,Zhang2021dive} in the form of the Nadaraya-Watson
kernel regression model \cite{Nadaraya-1964,Watson-1964}.

Let us consider a dataset $D=\{(\mathbf{x}_{1},y_{1}),(\mathbf{x}_{2}%
,y_{2}),...,(\mathbf{x}_{n},y_{n})\}$ consisting of $n$ instances, where
$\mathbf{x}_{i}=(x_{i1},...,x_{im})\in \mathbb{R}^{m}$ is a feature vector
involving $m$ features, $y_{i}\in \mathbb{R}$ represents the regression target
variable. The regression task is to learn a function $f:\mathbb{R}%
^{m}\rightarrow \mathbb{R}$ on the dataset $D$ such that the trained function
$f$ can predict the target value $\tilde{y}$ of a new observation $\mathbf{x}$.

According to the Nadaraya-Watson regression model
\cite{Nadaraya-1964,Watson-1964}, the target value $y$ for a new vector of
features $\mathbf{x}$ can be computed by using the weighted average of the
form:
\begin{equation}
\tilde{y}=f(\mathbf{x})=\sum_{i=1}^{n}\alpha(\mathbf{x},\mathbf{x}_{i})y_{i}.
\label{Expl_At_10}%
\end{equation}

Here weight $\alpha(\mathbf{x},\mathbf{x}_{i})$ indicates how close the vector
$\mathbf{x}_{i}$ from the dataset $D$ to the vector $\mathbf{x}$ that is the
closer the vector $\mathbf{x}_{i}$ to $\mathbf{x}$, the greater the weight
$\alpha(\mathbf{x},\mathbf{x}_{i})$ assigned to $y_{i}$.

The Nadaraya-Watson kernel regression uses kernels $K$ in order to express
weights $\alpha(\mathbf{x},\mathbf{x}_{i})$, in particular, the weights can be
computed as:
\begin{equation}
\alpha(\mathbf{x},\mathbf{x}_{i})=\frac{K(\mathbf{x},\mathbf{x}_{i})}%
{\sum_{j=1}^{n}K(\mathbf{x},\mathbf{x}_{j})}. \label{Expl_At_11}%
\end{equation}

If to apply terms introduced for the attention mechanism in
\cite{Bahdanau-etal-14}, then weights $\alpha(\mathbf{x},\mathbf{x}_{i})$ are
called as the attention weights, the target values $y_{i}$ are called values,
vectors $\mathbf{x}$ and $\mathbf{x}_{i}$ are called query and keys,
respectively. It should be noted that the original Nadaraya-Watson kernel
regression is a non-parametric model, i.e., it is an example of the
non-parametric attention pooling. However, the weights can be added by
trainable parameters which results the parametric attention pooling. In
particular, one of the well-known kernels in (\ref{Expl_At_11}) is the
Gaussian kernel which produces the softmax function of the Euclidean distance.
The attention weights with trainable parameters may have the form
\cite{Bahdanau-etal-14}:%
\begin{equation}
\alpha(\mathbf{x},\mathbf{x}_{i})=\text{\textrm{softmax}}\left(
\mathbf{q}^{\mathrm{T}}\mathbf{k}_{i}\right)  =\frac{\exp \left(
\mathbf{q}^{\mathrm{T}}\mathbf{k}_{i}\right)  }{\sum_{j=1}^{n}\exp \left(
\mathbf{q}^{\mathrm{T}}\mathbf{k}_{j}\right)  }, \label{Expl_At_12}%
\end{equation}
where $\mathbf{q=W}_{q}\mathbf{x}$, $\mathbf{k}_{i}\mathbf{=W}_{k}%
\mathbf{x}_{i}$, $\mathbf{W}_{q}$ and $\mathbf{W}_{k}$ are matrices of
trainable parameters.

Many definitions of attention weights and the attention mechanisms can be
presented, for example, the additive attention \cite{Bahdanau-etal-14},
multiplicative or dot-product attention \cite{Luong-etal-2015,Vaswani-etal-17}%
. A new attention mechanism is proposed below, which is based on training the
weighted GBMs and the Huber's $\epsilon$-contamination model.

\subsection{A brief introduction to the GBM for regression}

If to return to the regression problem stated above, then we aim to construct
a regression model or an approximation $g$ of the function $f$ that minimizes
the expected risk or the expected loss function
\begin{align}
L(g)  & =\mathbb{E}_{(\mathbf{x},y)\sim P}~L(y,g(\mathbf{x}))\nonumber \\
& =\int_{\mathcal{X}\times \mathbb{R}}L(y,g(\mathbf{x}))\mathrm{d}%
P(\mathbf{x},y),\label{Imp_SVM16}%
\end{align}
with respect to the function parameters. Here $P(\mathbf{x},y)$ is a joint
probability distribution of $\mathbf{x}$ and $y$; the loss function
$L(\cdot,\cdot)$ may be represented, for example, as follows:
\begin{equation}
L(y,g(\mathbf{x}))=\left(  y-g(\mathbf{x})\right)  ^{2}.\label{grad_bost_8}%
\end{equation}

Among many machine learning methods, which solve the regression problem, for
example, random forests \cite{Breiman-2001} and the support vector regression
\cite{Smola-Scholkopf-2004}), the GBM \cite{Friedman-2002} is one of the most
accurate methods.

Generally, GBMs iteratively improve the predictions of $y$ from $\mathbf{x}$
with respect to the loss function $L$. It is carried out by starting from an
approximation of $g$, for example, from some constant $c$, and then adding new
weak or base learners that improve upon the previous ones $M$ times. As a
result, an additive ensemble model of size $M$ is formed:
\begin{equation}
g_{0}(\mathbf{x})=c,\  \ g_{i}(\mathbf{x})=g_{i-1}(\mathbf{x})+\gamma_{i}%
h_{i}(\mathbf{x}),\ i=1,...,M. \label{grad_bost_10}%
\end{equation}
where $h_{i}$ is the $i$-th base model at the $i$-th iteration; $\gamma_{i}$
is the coefficient or the weight of the $i$-th base model.

Many GBMs use decision trees as the most popular base learners. The GBM is
represented in the form of Algorithm \ref{alg:Interpr_GBM_0}. It can be seen
from Algorithm \ref{alg:Interpr_GBM_0} that it minimizes the expected loss
function $L$ by computing the gradient iteratively. Each decision tree in the
GBM is constructed at each iteration to fit the negative gradients. The
function $h_{i}$ can be defined by parameters $\theta_{i}$, i.e.,
$h_{i}(\mathbf{x})=h(\mathbf{x},\theta_{i})$. It is trained on a new dataset
$\{(\mathbf{x}_{j},q_{j}^{(i)})\}$, where $q_{j}^{(i)}$, $j=1,...,n$, are
residuals defined as partial derivatives of the expected loss function at each
point $\mathbf{x}_{i}$ (see (\ref{Interpr_GBM_10})).

\begin{algorithm}
\caption{The original GBM algorithm} \label{alg:Interpr_GBM_0}
\begin{algorithmic}
[1]\REQUIRE Training set $D$; the number of iterations $T$
\ENSURE Prediction $g(\mathbf{x})$ for an instance $\mathbf{x}$
\STATE Initialize the function $g_{0}(\mathbf{x})=c$
\FOR{$t=1$, $t\leq T$ } \STATE Calculate the residual $q_{i}^{(t)}$ as the
partial derivative of the expected loss function $L(y_{i},g_{t}(\mathbf{x}%
_{i}))$ at each point of the training set:
\begin{equation}
q_{i}^{(t)}=-\left.  \frac{\partial L(y_{i},z)}{\partial z}\right \vert
_{z=g_{i-1}(\mathbf{x}_{i})},\ i=1,...,n \label{Interpr_GBM_10}%
\end{equation}
\STATE Train a base model $h_{t}(\mathbf{x}_{i})$ on a new dataset with
residuals $\{(\mathbf{x}_{i},q_{i}^{(t)})\}$
\STATE Find the best gradient descent step-size $\gamma_{t}$:%
\begin{equation}
\gamma_{t}=\arg \min_{\gamma}\sum_{i=1}^{n}L(y_{t},g_{t-1}(\mathbf{x}%
_{i})+\gamma h_{t}(\mathbf{x}_{i}))
\end{equation}
\STATE Update the function $g_{t}(\mathbf{x})=g_{t-1}(\mathbf{x})+\gamma
_{t}h_{t}(\mathbf{x})$
\ENDFOR
\STATE The resulting function after $T$ iterations is
\begin{equation}
g_{T}(\mathbf{x})=\sum_{t=1}^{T}\gamma_{t}h_{t}(\mathbf{x})=g_{T-1}%
(\mathbf{x})+\gamma_{T}h_{T}(\mathbf{x}).
\end{equation}
\end{algorithmic}
\end{algorithm}

\section{Attention-based GBM}

The idea to apply the attention mechanism to random forests was proposed in
\cite{Utkin-Konstantinov-22a}. Let us consider how this idea can be adapted to
the GBM that is how attention weights can be used in the GBM.

First, we consider the simplest case of the attention weights. This is a way
of the direct assignment of non-parametric weights to trees ($h_{t}$) without
trainable parameter. We also assume that the squared error loss function
(\ref{grad_bost_8}) is used. Then
\begin{align}
g_{T}(\mathbf{x})  &  =h_{0}(\mathbf{x})+\sum_{t=1}^{T}\gamma_{t}%
h_{t}(\mathbf{x})\nonumber \\
&  =h_{0}(\mathbf{x})+\sum_{t=1}^{T}\frac{1}{T}\left(  \gamma_{t}\cdot T\cdot
h_{t}(\mathbf{x})\right) \nonumber \\
&  =h_{0}(\mathbf{x})+\sum_{t=1}^{T}\omega_{t}\cdot \hat{h}_{t}(\mathbf{x}),
\label{GBM_Att_10}%
\end{align}
where $\omega_{t}=1/T$, $\hat{h}_{t}(\mathbf{x})=\gamma_{t}\cdot T\cdot
h_{t}(\mathbf{x})$ is the tree prediction $h_{t}(\mathbf{x})$ multiplied by
$\gamma_{t}$ and $T$.

It should be noted that weights $\omega_{i}$ can be generalized by taking any
values satisfying the weight conditions: $\sum_{t=1}^{T}\omega_{t}=1$ and
$\omega_{t}\geq0$, $t=1,...,T$. For convenience, we represent $\hat{h}_{t}$ as
a decision tree with the same structure as $h_{t}$, but with modified leaf values.

Let us consider now a decision tree as the weak learner in the GBM. Suppose
the set $\mathcal{J}_{i}^{(t)}$ represents indices of instances which 
fall into the $i$-th leaf after training the tree at the $t$-th iteration of
GBM. Define the mean vector $\mathbf{A}_{t}(\mathbf{x)}$ and the mean residual
value $B_{t}$ as the mean of training instance vectors, which fall into the
$i$-th leaf of a tree at the $t$-th iteration, and the corresponding observed
mean residual value at the same iteration, respectively, i.e.,
\begin{equation}
\mathbf{A}_{t}(\mathbf{x)}=\frac{1}{\# \mathcal{J}_{i}^{(t)}}\sum
_{j\in \mathcal{J}_{i}^{(t)}}\mathbf{x}_{j}, \label{RF_Att_20}%
\end{equation}%
\begin{equation}
B_{t}(\mathbf{x)}=\frac{1}{\# \mathcal{J}_{i}^{(t)}}\sum_{i\in \mathcal{J}%
_{j}^{(t)}}\hat{h}_{t}(\mathbf{x}_{j}). \label{RF_Att_21}%
\end{equation}

The distance between $\mathbf{x}$ and $\mathbf{A}_{t}(\mathbf{x)}$ indicates
how close the vector $\mathbf{x}$ to vectors $\mathbf{x}_{j}$ from the dataset
$D$ which fall into the same leaf as $\mathbf{x}$. We apply the $L_{2}$-norm
for the distance definition, i.e., there holds
\begin{equation}
d\left(  \mathbf{x},\mathbf{A}_{t}(\mathbf{x)}\right)  =\left \Vert
\mathbf{x}-\mathbf{A}_{t}(\mathbf{x)}\right \Vert ^{2}. \label{GBM_Att_15}%
\end{equation}

If we return to the Nadaraya-Watson regression model, then the GBM for
predicting the target value of $\mathbf{x}$ can be written in terms of the
regression model as:
\begin{equation}
G(\mathbf{x},\mathbf{w})=h_{0}(\mathbf{x})+\sum_{t=1}^{T}\alpha \left(
\mathbf{x},\mathbf{A}_{t}(\mathbf{x)},\mathbf{w}\right)  \cdot B_{t}%
(\mathbf{x)}. \label{RF_Att_47}%
\end{equation}

Here $\alpha \left(  \mathbf{x},\mathbf{A}_{t}(\mathbf{x)},\mathbf{w}\right)  $
is the attention weight which is defined in (\ref{Expl_At_10}) and depends on
the mean vector $\mathbf{A}_{t}(\mathbf{x)}$ and on the vector $\mathbf{w}$ of
training attention parameters. Here we can say that $B_{t}(\mathbf{x)}$ is the
value, $\mathbf{A}_{t}(\mathbf{x)}$ is the key, and $\mathbf{x}$ is the query
in terms of the attention mechanism. The weights satisfy the following
condition:
\begin{equation}
\sum_{t=1}^{T}\alpha \left(  \mathbf{x},\mathbf{A}_{t}(\mathbf{x)}%
,\mathbf{w}\right)  =1.
\end{equation}
The optimal parameters $\mathbf{w}$ can be found by minimizing the expected
loss function over a set $\mathcal{W}$ of parameters as follows:
\begin{equation}
\mathbf{w}_{opt}=\arg \min_{\mathbf{w}\in \mathcal{W}}~\sum_{j=1}^{n}\left(
y_{j}-G(\mathbf{x}_{j},\mathbf{w})\right)  . \label{RF_Att_49}%
\end{equation}

The next question is how to define the attention weights $\alpha \left(
\mathbf{x},\mathbf{A}_{t}(\mathbf{x)},\mathbf{w}\right)  $ such that the
optimization problem (\ref{RF_Att_49}) would be simply solved. An efficient
way for defining the attention weights has been proposed in
\cite{Utkin-Konstantinov-22a} where the well-known Huber's $\epsilon
$-contamination model \cite{Huber81} was used for trainable parameters
$\mathbf{w}$. The Huber's $\epsilon$-contamination model can be represented
as
\begin{equation}
(1-\epsilon)\cdot P+\epsilon \cdot Q, \label{GBM_Att_20}%
\end{equation}
where the probability distribution $P$ is contaminated by some arbitrary
distribution $Q$; the rate $\epsilon \in \lbrack0,1]$ is a model parameter
(contamination parameter) which reflects how \textquotedblleft
close\textquotedblright \ we feel that $Q$ must be to $P$ \cite{Berger85}.

If we assume that the attention weights $\alpha \left(  \mathbf{x}%
,\mathbf{A}_{t}(\mathbf{x)},\mathbf{w}\right)  $ are expressed through the
softmax function $\mathrm{softmax}(d\left(  \mathbf{x},\mathbf{A}%
_{t}(\mathbf{x)}\right)  )$ (see (\ref{Expl_At_12})), which provides the
probability distribution $P$ of the distance $d\left(  \mathbf{x}%
,\mathbf{A}_{t}(\mathbf{x)}\right)  $ between vector $\mathbf{x}$ and the mean
vector $\mathbf{A}_{t}(\mathbf{x)}$ with some parameters for all iterations,
$t=1,...,T$, then correction of the distribution can be carried out by means
of the $\epsilon$-contamination model, i.e., by means of the probability
distribution $Q$. Suppose that the distribution $Q$ is produced by parameters
$\mathbf{w}$ which can take arbitrary values in the $T$-dimensional unit
simplex $\mathcal{W}$, i.e., $Q$ is arbitrary such that $w_{1}+...+w_{T}=1$.
Then we can write the attention weights as (\ref{GBM_Att_20}), i.e.,
\begin{align}
&  \alpha \left(  \mathbf{x},\mathbf{A}_{t}(\mathbf{x)},\mathbf{w}\right)
\nonumber \\
&  =(1-\epsilon)\cdot P+\epsilon \cdot Q\nonumber \\
&  =(1-\epsilon)\cdot \mathrm{softmax}(d\left(  \mathbf{x},\mathbf{A}%
_{t}(\mathbf{x)}\right)  )+\epsilon \cdot w_{t}.\label{GBM_Att_30}%
\end{align}

The main advantage of the above representation is that the attention weights
linearly depend on the trainable parameters $\mathbf{w}$. We will see below
that this representation leads to the quadratic optimization problem for
computing optimal weights. This is a very important property of the proposed
attention weights because we do not need to numerically solve complex
optimization problem for computing the weights. The standard quadratic
optimization problem can be only solved, which has a unique solution. The
parameter $\epsilon$ is the tuning parameter. It is changed from $0$ to $1$ to
get some optimal value which allows us to get the highest regression
performance on the validation set. Moreover, the attention weights use the
non-trainable softmax function which is simply computed and does not need to
be learned.

Substituting (\ref{GBM_Att_15}) into (\ref{GBM_Att_30}), we get the following
final form of the attention weights:
\begin{align}
&  \alpha \left(  \mathbf{x},\mathbf{A}_{t}(\mathbf{x)},\mathbf{w}\right)
\nonumber \\
&  =(1-\epsilon)\cdot \mathrm{softmax}\left(  \frac{\left \Vert \mathbf{x}%
-\mathbf{A}_{t}(\mathbf{x)}\right \Vert ^{2}}{2}\delta^{t}\right)
+\epsilon \cdot w_{t}.
\end{align}

Here $\delta \in \lbrack0,1]$ is the discount factor such that $\delta^{t}$ is
decreased with the number of iteration $t$. It determines how much the impact
of the attention weight is decreased with the number of iteration $t$. The
discount factor is a tuning parameter.

Hence, the expected loss function for training parameters $w$ can be written
as:
\begin{equation}
\min_{\mathbf{w\in}\mathcal{W}}\sum_{s=1}^{n}\left(  y_{s}-h_{0}%
(\mathbf{x})-\sum_{t=1}^{T}F_{t}(\mathbf{x}_{s},\delta^{t},\epsilon
,w_{t})\right)  ^{2},\label{GBM_Att_42}%
\end{equation}
where
\begin{equation}
F_{t}(\mathbf{x},\delta^{t},\epsilon,w_{t})=B_{t}(\mathbf{x)}\left(
(1-\epsilon)D_{t}(\mathbf{x},\delta^{t})+\epsilon \cdot w_{t}\right)  ,
\end{equation}
\begin{equation}
D_{t}(\mathbf{x},\delta^{t})=\text{\textrm{softmax}}\left(  \frac{\left \Vert
\mathbf{x}-\mathbf{A}_{t}(\mathbf{x)}\right \Vert ^{2}}{2}\delta^{t}\right)  ,
\end{equation}

Problem (\ref{GBM_Att_42}) is the standard quadratic optimization problem with
linear constraints $\mathbf{w}\in \mathcal{W}$, i.e., $w_{t}\geq0$,
$t=1,...,T$, and $\sum_{t=1}^{T}w_{t}=1$.

As a result, we get a simple quadratic optimization problem whose solution
does not meet any difficulties. from the computational point of view because
its training is based on solving the standard quadratic optimization problem.

\section{Numerical experiments}

The attention-based GBM is evaluated and investigated solving regression
problems on $11$ datasets from open sources. Dataset Diabetes can be found in
the corresponding R Packages. Three datasets Friedman 1, 2 3 are described at
site: https://www.stat.berkeley.edu/\symbol{126}breiman/bagging.pdf. Datasets
Regression and Sparse are available in package \textquotedblleft
Scikit-Learn\textquotedblright. Datasets Wine Red, Boston Housing, Concrete,
Yacht Hydrodynamics, Airfoil are taken from UCI Machine Learning Repository
\cite{Dua:2019}. A brief introduction about these data sets is represented in
Table \ref{t:regres_datasets} where $m$ and $n$ are numbers of features and
instances, respectively. A more detailed information is available from the
above resources.%

%TCIMACRO{\TeXButton{B}{\begin{table}[tbp] \centering}}%
%BeginExpansion
\begin{table}[h] \centering
%EndExpansion
\caption{A brief introduction about the regression data sets}%
\begin{tabular}
[c]{cccc}\hline
Data set & Abbreviation & $m$ & $n$\\ \hline
Diabetes & Diabetes & $10$ & $442$\\ \hline
Friedman 1 & Friedman 1 & $10$ & $100$\\ \hline
Friedman 2 & Friedman 2 & $4$ & $100$\\ \hline
Friedman 3 & Friedman 3 & $4$ & $100$\\ \hline
Scikit-Learn Regression & Regression & $100$ & $100$\\ \hline
Scikit-Learn Sparse Uncorrelated & Sparse & $10$ & $100$\\ \hline
UCI Wine red & Wine & $11$ & $1599$\\ \hline
UCI Boston Housing & Boston & $13$ & $506$\\ \hline
UCI Concrete & Concrete & $8$ & $1030$\\ \hline
UCI Yacht Hydrodynamics & Yacht & $6$ & $308$\\ \hline
UCI Airfoil & Airfoil & $5$ & $1503$\\ \hline
\end{tabular}
\label{t:regres_datasets}%
%TCIMACRO{\TeXButton{E}{\end{table}}}%
%BeginExpansion
\end{table}%
%EndExpansion

We use the coefficient of determination denoted $R^{2}$ and the mean absolute
error (MAE) for the regression evaluation. The greater the value of the
coefficient of determination and the smaller the MAE, the better results we
get. Every GBM has $200$ iterations. Decision trees at each iteration are
built such that at least $10$ instances fall into every leaf of trees. This
condition is used to get desirable estimates of vectors $\mathbf{A}%
_{t}(\mathbf{x}_{s}\mathbf{)}$.

To evaluate the average accuracy measures, we perform a cross-validation with
$100$ repetitions, where in each run, we randomly select $n_{\text{tr}}=4n/5$
training data and $n_{\text{test}}=n/5$ testing data. The best results in all
tables are shown in bold.

In all tables, we compare $R^{2}$ and the MAE for three cases: (\textbf{GBM})
the GBM without the softmax and without attention model; (\textbf{Non-param})
a special case of the AGBoost model when trainable parameters $\mathbf{w}$ are
not learned, and they are equal to $1/T$; (\textbf{AGBoost}) the proposed
AGBoost model with trainable parameters $\mathbf{w}$.

The optimal values of the contamination parameter $\epsilon_{opt}$ and the
discount factor $\delta_{opt}$ are provided. The case $\epsilon_{opt}=1$ means
that the attention weights are totally determined by the tree weights and do
not depend on each instance. The case $\epsilon_{opt}=0$ means that weights of
trees are determined only by the softmax function without trainable parameters.%

%TCIMACRO{\TeXButton{B}{\begin{table}[tbp] \centering}}%
%BeginExpansion
\begin{table*}[h] \centering
%EndExpansion
\caption{Measures $R^2$ and MAE for comparison of three models by optimal values of contamination parameter and discount factor with the original decision trees as base learners}%
\begin{tabular}
[c]{ccccccccc}\hline
&  &  & \multicolumn{3}{c}{$R^{2}$} & \multicolumn{3}{c}{MAE}\\ \hline
Data set & $\epsilon_{opt}$ & $\delta_{opt}$ & GBM & Non-param & AGBoost &
GBM & Non-param & AGBoost\\ \hline
Diabetes & $0$ & $0.9$ & $0.390$ & $\mathbf{0.391}$ & $\mathbf{0.391}$ &
$\mathbf{46.11}$ & $46.53$ & $46.53$\\ \hline
Friedman 1 & $1$ & $0.01$ & $0.582$ & $0.582$ & $\mathbf{0.595}$ & $2.269$ &
$2.269$ & $\mathbf{2.219}$\\ \hline
Friedman 2 & $0.778$ & $1$ & $0.909$ & $0.913$ & $\mathbf{0.944}$ & $85.39$ &
$83.79$ & $\mathbf{62.21}$\\ \hline
Friedman 3 & $0.667$ & $0.9$ & $0.696$ & $0.700$ & $\mathbf{0.715}$ & $0.135$
& $0.136$ & $\mathbf{0.125}$\\ \hline
Regression & $0.778$ & $1$ & $0.572$ & $0.577$ & $\mathbf{0.626}$ & $88.34$ &
$86.91$ & $\mathbf{82.00}$\\ \hline
Sparse & $0.000$ & $1$ & $0.598$ & $\mathbf{0.651}$ & $\mathbf{0.651}$ &
$1.646$ & $\mathbf{1.582}$ & $\mathbf{1.582}$\\ \hline
Airfoil & $0.889$ & $1$ & $0.792$ & $0.794$ & $\mathbf{0.831}$ & $2.456$ &
$2.443$ & $\mathbf{2.156}$\\ \hline
Boston & $0.667$ & $1$ & $0.832$ & $0.838$ & $\mathbf{0.845}$ & $2.481$ &
$2.465$ & $\mathbf{2.379}$\\ \hline
Concrete & $0.889$ & $1$ & $0.847$ & $0.846$ & $\mathbf{0.872}$ & $4.970$ &
$4.981$ & $\mathbf{4.382}$\\ \hline
Wine & $0.111$ & $0.5$ & $0.412$ & $0.412$ & $\mathbf{0.413}$ & $0.475$ &
$0.475$ & $\mathbf{0.470}$\\ \hline
Yacht & $1$ & $0.01$ & $0.973$ & $0.973$ & $\mathbf{0.991}$ & $1.594$ &
$1.594$ & $\mathbf{0.675}$\\ \hline
\end{tabular}
\label{t:GBM_regression_0}%
%TCIMACRO{\TeXButton{E}{\end{table}}}%
%BeginExpansion
\end{table*}%
%EndExpansion

Measures $R^{2}$ and MAE for three cases (GBM, Non-param and AGBoost) are
shown in Table \ref{t:GBM_regression_0} with the original decision trees as
base learners. It can be seen from Table \ref{t:GBM_regression_0} that the
proposed AGBoost model outperforms the GBM itself and the non-parametric model
or is comparable with these models for all datasets.

To formally show the outperformance of the proposed AGBoost model with the
original decision trees as base learners, we apply the $t$-test which has been
proposed and described by Demsar \cite{Demsar-2006} for testing whether the
average difference in the performance of two models, AGBoost and GBM, is
significantly different from zero. Since we use differences between accuracy
measures of AGBoost and GBM, then they are compared with $0$. The $t$
statistics in this case is distributed according to the Student distribution
with $11-1$ degrees of freedom. Results of computing the $t$ statistics of the
difference are the p-values denoted as $p$ and the $95\%$ confidence interval
for the mean $0.024$, which are $p=0.0013$ and $[0.012,0.037]$, respectively.
The $t$-test demonstrates the outperformance of AGBoost in comparison with the
GBM because $p<0.05$. We also compare AGBoost with the non-parametric model.
We get the $95\%$ confidence interval for the mean $0.018$, which are
$p=0.0045$ and $[0.007,0.029]$, respectively. Results of the second test also
demonstrate the outperformance of AGBoost in comparison with the
non-parametric model.%

%TCIMACRO{\TeXButton{B}{\begin{table}[tbp] \centering}}%
%BeginExpansion
\begin{table*}[h] \centering
%EndExpansion
\caption{Measures $R^2$ and MAE for comparison of three models by optimal values of contamination parameter and discount factor with the ERT as base learners}%
\begin{tabular}
[c]{ccccccccc}\hline
&  &  & \multicolumn{3}{c}{$R^{2}$} & \multicolumn{3}{c}{MAE}\\ \hline
Data set & $\epsilon_{opt}$ & $\delta_{opt}$ & GBM & Non-param & AGBoost &
GBM & Non-param & AGBoost\\ \hline
Diabetes & $0.111$ & $0.01$ & $0.435$ & $0.434$ & $\mathbf{0.440}$ & $44.80$ &
$44.83$ & $\mathbf{44.11}$\\ \hline
Friedman 1 & $1$ & $0.01$ & $0.561$ & $0.561$ & $\mathbf{0.612}$ & $2.248$ &
$2.248$ & $\mathbf{2.146}$\\ \hline
Friedman 2 & $1$ & $0.01$ & $0.874$ & $0.874$ & $\mathbf{0.975}$ & $98.57$ &
$98.57$ & $\mathbf{41.12}$\\ \hline
Friedman 3 & $0.667$ & $1$ & $0.637$ & $0.639$ & $\mathbf{0.800}$ & $0.160$ &
$0.158$ & $\mathbf{0.110}$\\ \hline
Regression & $0.778$ & $0.9$ & $0.495$ & $0.487$ & $\mathbf{0.604}$ & $96.88$
& $97.76$ & $\mathbf{83.30}$\\ \hline
Sparse & $0.556$ & $1$ & $0.558$ & $0.611$ & $\mathbf{0.659}$ & $1.777$ &
$1.656$ & $\mathbf{1.504}$\\ \hline
Airfoil & $1$ & $0.01$ & $0.747$ & $0.747$ & $\mathbf{0.835}$ & $2.734$ &
$2.734$ & $\mathbf{2.151}$\\ \hline
Boston & $0.778$ & $1$ & $0.835$ & $0.837$ & $\mathbf{0.868}$ & $2.463$ &
$2.457$ & $\mathbf{2.260}$\\ \hline
Concrete & $1$ & $0.01$ & $0.817$ & $0.817$ & $\mathbf{0.867}$ & $5.577$ &
$5.577$ & $\mathbf{4.565}$\\ \hline
Wine & $0.778$ & $1$ & $0.396$ & $0.400$ & $\mathbf{0.409}$ & $0.484$ &
$0.481$ & $\mathbf{0.468}$\\ \hline
Yacht & $1$ & $0.01$ & $0.973$ & $0.973$ & $\mathbf{0.992}$ & $1.596$ &
$1.596$ & $\mathbf{0.642}$\\ \hline
\end{tabular}
\label{t:GBM_regression_1}%
%TCIMACRO{\TeXButton{E}{\end{table}}}%
%BeginExpansion
\end{table*}%
%EndExpansion

Measures $R^{2}$ and MAE for three cases (GBM, Non-param and AGBoost) are
shown in Table \ref{t:GBM_regression_1} with the ERT as base learners. It can
be seen from Table \ref{t:GBM_regression_1} that the proposed AGBoost model
outperforms the GBM itself and the non-parametric model for all datasets.

To formally show the outperformance of the proposed AGBoost model with ERTs as
base learners, we again apply the $t$-test. Results of computing the $t$
statistics are the p-values and the $95\%$ confidence interval for the mean
$0.067$, which are $p=0.0012$ and $[0.033,0.100]$, respectively. The $t$-test
demonstrates the clear outperformance of AGBoost in comparison with the GBM.
We also compare AGBoost with the non-parametric model. We get the $95\%$
confidence interval for the mean $0.062$, which are $p=0.0019$ and
$[0.029,0.095]$, respectively. Results of the second test also demonstrate the
outperformance of AGBoost in comparison with the non-parametric model. It
follows from the obtained results that models with the ERTs as base learners
provide better results than the models with original decision trees.

\section{Concluding remarks}

A new model of the attention-based GBM has been proposed. It can be regarded
as an extension of the attention-based random forest
\cite{Utkin-Konstantinov-22a}. The proposed model inherits advantages of the
attention mechanism and the GBM. Moreover, it allows us to avoid using neural
networks. Numerical experiments have demonstrated that incorporating the
attention model into the GBM improves the original GBM.

At the same time, the proposed model is rather flexible and this fact allows
us to determine several directions for further research. First, we have
investigated only the Huber's $\epsilon$-contamination model for incorporating
the trainable parameter into the attention. However, there exist some
statistical models which have similar properties. Their use and study instead
of the Huber's contamination model is a direction for further research. The
proposed AGBoost model uses non-parametric softmax function for computing the
attention weights. It is interesting to extend the proposed model to the case
of the parametric softmax function with trainable parameters. It should be
noted that additional trainable parameters in the softmax may significantly
complicate the model. However, efficient computation algorithms are also
directions for further research. It should be also noted that the proposed
attention-based approach can be incorporated into other GBM models, for
example, into XGBoost, pGBRT, SGB, etc. This is also a direction for further research.

\section*{Acknowledgement}

This work is supported by the Russian Science Foundation under grant 21-11-00116.

\bibliographystyle{IEEEtran}
\bibliography{Attention,Boosting,Classif_bib,Deep_Forest,Expl_Attention,Explain,Explain_med,Imprbib,Lasso,MIL,MYBIB,MYUSE}

\end{document}